%% file: main.tex
\documentclass[10pt,twocolumn,a4paper]{article}

\usepackage{titling}

 \usepackage{algorithm}
 \usepackage{algpseudocode}
\usepackage{amsmath} 
\usepackage{authblk}
 \usepackage{color}
 \usepackage{comment}
 \usepackage{enumitem}
\usepackage{graphicx} 
\usepackage{latexsym}
 \usepackage{mdframed}
\usepackage{multirow} 
\usepackage{microtype}
\usepackage{subcaption} 
 \usepackage{soul} 
\usepackage{times}
\usepackage[hyphens]{url}
 \usepackage{verbatim} 
\usepackage{wrapfig}

\DeclareMathOperator*{\argmax}{arg\,max}

\title{Modeling Dynamic Relationships Between Characters in Literary Novels}

\author[1]{Snigdha Chaturvedi}
\author[2]{Shashank Srivastava}
\author[1]{Hal Daum\'{e} III}
\author[2]{Chris Dyer}
\affil[1]{Department of Computer Science, University of Maryland, College Park}
\affil[2]{Department of Computer Science, Carnegie Mellon University}

\date{}

\begin{document}
\maketitle
\begin{abstract}
Studying characters plays a vital role in computationally representing and interpreting narratives. Unlike previous work, which has focused on inferring character roles, we focus on the problem of modeling their relationships. Rather than assuming a fixed relationship for a character pair, we hypothesize that relationships are dynamic and temporally evolve with the progress of the narrative, and formulate the problem of relationship modeling as a structured prediction problem. We propose a semi-supervised framework to learn \emph{relationship sequences} from fully as well as partially labeled data. We present a Markovian model capable of accumulating historical beliefs about the relationship and status changes. We  use a set of rich linguistic and semantically motivated features that incorporate world knowledge to investigate the textual content of narrative. We empirically demonstrate that such a framework outperforms competitive baselines. 

\end{abstract}

\section{Introduction}
\input{introduction}

\section{Relationship Prediction Model}
\input{models}

\section{Feature Engineering}
\input{features}

\section{Empirical Evaluation}
\input{experiments}


\section{Literature Survey}
\input{relatedWork}

\section{Conclusion and Discussion}
\input{conclusion}

\bibliographystyle{acl}
\bibliography{main}

\end{document}

%% file: introduction.tex
\label{introduction}

\begin{center}
\begin{figure}[tb]
\centering
\includegraphics[width=\linewidth]{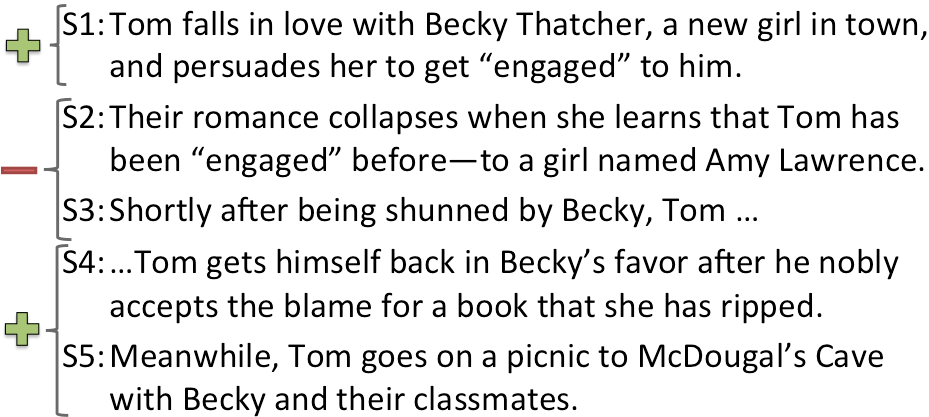}
\caption{
Sample sentences from a narrative depicting evolving relationship between characters: Tom and Becky. The relationship changes from cooperative (+) to non-cooperative (-) and then back to cooperative (+). `\ldots' represent text omitted due to space constraints.
}
\label{fig:introEg}
\end{figure}
\end{center}

\vspace{-0.5cm}
The field of computational narrative studies focuses on algorithmically understanding, representing and generating stories. Most research in this field focuses on modeling the narrative from the perspective of (1) events or (2) characters. 


Popular events-based approaches include scripts~\cite{Schank:1977,Regneri:2010}, plot units~\cite{goyal2010automatically,McIntyre:2010,finlayson2012learning,Elsner:2012},  temporal event chains or schemas~\cite{Chambers:2008,Chambers:2009}, and the more recent bags of related events~\cite{Orr:2014,Chambers:2013,Cheung:2013}.

The alternate perspective attempts to understand 
stories from the viewpoint of characters and relationships between them. This perspective explains the set of observed actions using characters' \emph{personas} or \emph{roles} and the expected behavior of the character in that role~\cite{Vargas:2014,Bamman:2013,Bamman:2014,Elson:2012}. Recent work has also focused on constructing social networks including signed social networks to model the relationships between individual characters~\cite{Agarwal:2013,Elson:2010,Krishnan:2015}.

The work presented in this paper aligns with 
the second perspective 
. We address the problem of modeling relationships between characters in literary fiction, specifically novels. Existing work mentioned above, 
models each character as assuming a single narrative role, and these roles define the relationships between characters and also govern their actions. While such a simplified assumption provides a good general overview of the narrative, it is not sufficient to explain \emph{all} events in the narrative. We believe that in most narratives, relationships between characters are not static but \emph{evolve} as the novel progresses.  For example, consider the relationship between \emph{Tom} and \emph{Becky} depicted in Fig.~\ref{fig:introEg} which shows an excerpt from the summary~\footnote{SparkNotes Editors. “SparkNote on The Adventures of Tom Sawyer.” SparkNotes LLC. 2003. \url{http://www.sparknotes.com/lit/tomsawyer/}} of \emph{The Adventures of Tom Sawyer} 
. For most of the narrative (and its summary), the characters are participants in a romantic relationship, which explains most, but not all, of their mutual behavior. 
However, we can observe that their relationship was not static but evolving, driving nature of the characters' actions. In this particular case, the characters presumably start as lovers (sentence S1 in the Figure), which is hinted at by (and explains) becoming engaged. The relationship sours when Tom reveals his previous love interest (S2 and S3). However, later in the narrative they reconcile (S4 and S5). 
A model that assumes a fixed romantic relationship between characters would fail to explain their behaviors during the phase when their relationship was under stress. 

Therefore, we assume that the relationship between characters evolves with the progress of the novel and model it as a sequence of latent variables denoting relation state. In this work, we take a coarse-grained view, and model relation states as binary variables (roughly indicating \emph{cooperative}/\emph{non-cooperative} relation at different points in the narrative). For instance 
in Fig.~\ref{fig:introEg}, the relationship between Tom and Becky can be represented by the sequence $\langle$cooperative, non-cooperative, cooperative$\rangle$. Given a narrative and a pair of characters appearing in it, we address the task of \textbf{learning relationship sequences.} The narrative fragment of interest for us 
is represented by the set of sentences in which the two characters of interest appeared together, arranged in the order of occurrence in the narrative. 

To address this problem we propose a semi-supervised segmentation framework for training on a collection of fully labeled and partially labeled  
sequences of sentences from narrative stories. The structured prediction model in the proposed framework attempts to  model the `narrative flow' in the sequence of sentences. Following 
previous work~\cite{propp1968,Bamman:2014}, it incorporates the linguistic and semantic information present in the sentences by tracking events and states associated with the characters of interest and enhances them with world knowledge~\cite{connotation:2013,Liu:2005,Wilson:2005}. We demonstrate the strength of our structured model by comparing it against an unstructured baseline that treats individual sentences independently. 

Our main contributions are as follows:
\begin{itemize}[leftmargin=*,noitemsep,nolistsep]
\itemsep0em 
\item {We formulate the novel problem of relationship modeling in narrative text as a structured prediction task instead of a categorical binary or multi-class classification problem.}
\item {We propose rich linguistic features that incorporate semantic and world knowledge.}
\item {We present a semi-supervised framework to incorporate the narrative structure of the text and empirically demonstrate that it outperforms competitive baselines.}
\end{itemize}

%% file: models.tex
\label{model}

In this section we describe our relationship-modeling framework in detail. Given the narrative text in form of a sequence of sentences (in which the two characters of interest appear together), $\mathbf{x}=\langle x_{1},x_{2},\ldots,x_{l}\rangle$, we address the problem of segmenting it into non-overlapping and semantically meaningful segments that represent continuities in relationship status. Each segment is labeled with a single relationship status $r_{j} \in \{-1,+1\}$ hence yielding a relationship sequence $\mathbf{r}=\langle r_{1},r_{2},\ldots,r_{k}\rangle k\leq l$. Our approach uses a second order Markovian latent variable model for segmentation that is embedded in semi-supervised framework to utilize varying levels of labeling in the data. 
We now describe our segmentation model and the semi-supervised framework in detail.

\subsection{Segmentation Model}
\label{segmentationModel}
This model forms the core of our framework. It assumes that each sentence in the sequence is associated with a latent state that represents its relationship status. 
While making this assignment, it analyzes the content of individual sentences using  a rich feature set  and simultaneously models the flow of information between the states
by treating the prediction task as a structured problem. 
We utilize a second-order Markov model that can remember a long history of the relationship between the two characters and collectively maximizes the following linear scores for individual sequences:
 \begin{equation}
\text{score} = \sum_i\mathbf{w} \boldsymbol{\phi}(\textbf{x}, y_{i}, y_{i-1}, y_{i-2})]
\label{eqn:svm1}
\end{equation}
where $x$ is the input sequence and $y_i$ denotes the latent state assignment of its $i^{th}$ sentence to a relationship segment. Individual $y_i$s collectively yield the relationship sequence, $\mathbf{r}$ (by collapsing consecutive  occurrences of identical states). $\boldsymbol{\phi}$ represents features at the $i^{th}$ sentence that depend on the current state, $y_i$, and the previous two states, $y_{i-1}$ and $y_{i-2}$, and $\mathbf{w}$ represents their weights. The second order Markov assumption of our features ensures continuity and coherence of behavior of the two characters within individual relationship segments. 

The linear segmentation model proposed here is trained using an averaged structured perceptron~\cite{Collins:2002}. For inference, it uses a Viterbi based dynamic programming algorithm. The extension of Viterbi to incorporate second order constraints is straightforward. We replace the reference to a state 
(in the state space $|Y|$) by a reference to a state pair (in the two fold product space $|Y|\times|Y|$).
Note that this precludes certain transitions while computing the Viterbi matrix, viz.: if the state pair at any point in narrative, $t$, is of the form $(s_i, s_j)$, then the set of state pair candidates at $t+1$ only consists of pairs of the form  $(s_j,s_k)$. Incorporating these constraints, we compute the Viterbi matrix and obtain the highest scoring state sequence by backtracking as usual. 

\subsection{Semi-supervised Framework}
\label{framework}
The above segmentation model requires labeled ($\mathbf{x},\mathbf{y}$) for training. However, given the nature of the task, we acknowledge that obtaining a huge dataset of labeled sequence can be time consuming as well as expensive. On the other hand, it might be more convenient to obtain partially labeled data especially in cases in which only a subset of the sentences of a sequence have an obvious relationship state membership. 
We, therefore, propose a semi-supervised framework, which can leverage partial supervision for training the segmentation model. This framework assumes that the training dataset consists of two types of labeled sequences: fully labeled, in which the complete state sequence is observed ${y_{i}} \forall i\in\{1\ldots l\}$ and partially labeled, in which some of the sentences of the sequence are annotated with ${y_{i}}$ such that $i\subset\{1\ldots l\}$. 

This framework uses a two step algorithm (Algorithm~\ref{fig:trainingAlgo}) to iteratively refine feature weights, $\mathbf{w}$, of the  segmentation model. In the first step, it uses existing weights, $\mathbf{w_n}$, to assign state sequences to the partially labeled instances. 
For state assignment we use a constrained version of the Viterbi algorithm that obtains the best possible state sequence that agrees with the partial ground truth. In other words, for the annotated sentences of a partially annotated sequence, it precludes all state assignments except the given ground truth, but segments the rest of the sequence optimally under these constraints. In the second step, we train the structured perceptron based segmentation model, using the ground truth and the state assignments obtained in the previous step, to obtain the refined weights $w_{n+1}$. Similar approaches have been used in the past~\cite{Srivastava:2014}.

\begin{algorithm}[ht!]
\caption{Training algorithm for the semi-supervised framework}
\begin{algorithmic}[1]
\State\textbf{Input:} Fully labeled, $F$ and partially labeled $P$ sequences\\
$T$: number of iterations
\State\textbf{Output:} Weights $\mathbf{w}$
\State\textbf{Initialization:} Initialize $\mathbf{w}$ randomly
\For {$n:1$ to $N$}
\State $\hat{\mathbf{y}}_j = \argmax_{\mathbf{y}_j} [\mathbf{w_n} \cdot \boldsymbol{\phi}(\mathbf{x}, \mathbf{y})_j]$   $\forall j \in P$ and $U$ such that $\hat{\mathbf{y}}_j$ agrees with the partial annotated states (ground truth). 
\State $w_{n+1}$ = AveragedStructuredPerceptron($\{(\mathbf{x}, \hat{\mathbf{y}})_j\}$ $\forall j \in \{P, F\}$)
\EndFor
\State return $w$ 
\end{algorithmic}
\label{fig:trainingAlgo}
\end{algorithm}

%% file: features.tex
\label{features}
We now describe the features used by our segmentation model. We first pre-processed the text of various novel summaries to obtain part-of-speech tags and dependency parses, identify major characters and perform character names clustering (assemble `Tom', `Tom Sawyer' etc.) using the Book-nlp pipeline~\cite{Bamman:2014}. However, the pipeline, designed for long text documents involving multiple characters, was slightly conservative while resolving co-references. We augmented its output using coreferences obtained from the Stanford Core NLP system~\cite{corenlp}. We also obtained a frame-semantic parse of the text using Semafor~\cite{Das:2014}.

After pre-processing, given two characters and a sequence of pre-processed sentences in which the two appeared together, we extracted the following features for individual sentences. 

\subsection{Content features}
These features help the model in characterizing the textual content of the sentences. They are based on the following general template which depends on the sentence, $x_j$, and its state, $y_j$:
$\phi(x_j,y_j) = \alpha$ if the current state is $y_j$ ; $0$ otherwise
where, $\alpha \in$ F1 to F33, where F1 to F33 are defined below.

\vspace{0.2cm}
\noindent \textbf{1. Actions based:} These features are motivated by Vladimir Propp's Structuralist narrative theory~\cite{propp1968} based insight that characters have a `sphere of actions'. We model the actions affecting the two characters by identifying all verbs in the sentence, their agents (using `nsubj' and `agent' dependency relations) and their patients (using `dobj' and `nsubjpass' relations). This information was extended using verbs conjunct to each other using `conj'. We also used the `neg' relation to determine the negation status of each verb. Based on this information we extracted the following features:
\begin{itemize}[leftmargin=*,noitemsep,nolistsep]
\itemsep0em 
\item {\textit{Are Team [F1]:} This feature models whether the two characters acted as a team. It is a binary feature indicating if the two characters were agents (or patients) of a verb together. }

\item {\textit{Acts Together [F2-F7]:} These features explicitly model the behavior of the two characters towards each other using verbs for which one of the characters was the agent and the other was patient. These six numeric features look at positive/negative connotation~\cite{connotation:2013}, sentiment~\cite{Liu:2005} and prior-polarity~\cite{Wilson:2005} of the verbs (while considering their negation status). 
}

\item {\textit{Surrogate Acts Together [F8-F13]:} The above features are designed to be high-precision features that directly analyze the nature of actions of the characters towards each other. However, their recall might suffer from limitations of the NLP pre-processing pipeline. For instance, a character might be an implicit/subtle patient of an action done by the other character. 
For example, Tom is not the direct patient of \emph{shunned} in S3 in Fig.~\ref{fig:introEg}. To include such cases we define a set of six surrogate features that, like before, consider positive and negative connotations, sentiments and prior-polarities of verbs (while considering negation). However, only those verbs are considered which have one of the characters as either the agent or the patient, and occur in sentences that did not contain any other character apart from the two of interest.}
\end{itemize}

\vspace{0.2cm}
\noindent \textbf{2. Adverb based:} These features model narrator's bias in describing characters' actions by analyzing the adverbs modifying the verbs identified in `Action based' features (using `advmod' dependency relations). For example, in S4 in Fig.~\ref{fig:introEg} the fact that Tom \emph{nobly} accepts the blame provides an evidence of a  positive relationship. 

\begin{itemize}[leftmargin=*,noitemsep,nolistsep]
\itemsep0em 
\item {\textit{Adverbs Together [F14-F19]} and \textit{Surrogate Adverbs Together [F20-F25]:} Six numeric features each measuring polarity of adverbs modifying the verbs considered in `Acts Together' and `Surrogate Acts Together' respectively.}


\end{itemize}

\vspace{0.2cm}
\noindent \textbf{3. Lexical [F26-27]:} These bag-of-words style features analyze the connotations of all words (excluding stop-words) occurring between pairs of mentions of the two characters in the sentence. E.g. in S5 in Fig.~\ref{fig:introEg} there is one pair of mentions of the two characters: $\langle$ Tom, Becky$\rangle$, and the words occurring between the two mentions are ``goes on a picnic to McDougal's cave with'' (stopwords included for readability).

\vspace{0.2cm}
\noindent \textbf{4. Semantic Parse based:} These features incorporate information from a framenet-style semantic parse of the sentence. To design these features, we manually compiled lists of \emph{frames} (along with corresponding relevant \emph{frame-elements}) with positive (or negative) connotations depending on whether they are indicative of positive (or negative) relationship between participants (identified in the corresponding frame-elements). 
Our set of lists also consisted of ambiguous frames like `cause\_bodily\_experience' in which case the exact connotation of the frame was determined on-the-fly depending on the lexical unit at which that frame fired. Lastly, we had a list of `Relationship' frames that indicated familial or professional relationship between participants. Table~\ref{table:framesList} shows examples of various types of frames and their relevant frame-elements. The complete list is available on the first author's webpage. Based on these lists, we extracted the following two types of features:

\begin{itemize}[leftmargin=*,noitemsep,nolistsep]
\itemsep0em 
\item {\textit{Frames Fired [F28-F30]:} Three numeric features counting number of positive, negative and `relationship' frames fired such that at least one of the characters belonged to the relevant frame-element.}

\item {\textit{Frames Fired [F31-F33]:} Three features counting number of positive, negative and `relationship' frames fired.}
\end{itemize}

\begin{table}
\begin{center}
\begin{tabular}{|p{0.22\columnwidth}|p{0.29\columnwidth}|p{0.33\columnwidth}|}\hline
\textbf{Type} & \textbf{Frame} & \textbf{Frame-elements}\\\hline
Negative        & `killing'                 & `killer', `victim'\\\cline{2-3}
                & `attack'                  & `assailant', `victim'\\\hline
Positive        & `forgiveness'             & `judge', `evaluee'\\\cline{2-3}
                & `supporting'              & `supporter', `supported'\\\hline
Ambiguous       & `cause bodily experience' & `agent', `experiencer'\\\cline{2-3}
                & `friendly or hostile'     & `side\_1', `side\_2', `sides'\\\hline
Relationship    & `kinship'                 & `alter', `ego', `relatives'\\\cline{2-3}
                & `subordinates and superiors' &`superior', `subordinate'\\\hline
\end{tabular}
\caption{Samples of various types of Frame-net frames used by `Semantic Parse based' features.}
\label{table:framesList}
\end{center}
\vspace{-0.1in}
\end{table} 

\subsection{Transition features} 
While content features assist the model in analyzing the text of individual sentences, these features enable the model to remember relationship histories, thus 
discouraging it from changing relationship states too frequently within a sequence.
\begin{itemize}[leftmargin=*,noitemsep,nolistsep]
\itemsep0em 
\item $\phi(y_j,y_{j-1},y_{j-2}) = 1$ if current state is $y_j$ and the previous two states were $y_{j-1},y_{j-2}$; $0$ otherwise
\item $\phi(y_j,y_{j-1}) = 1$ if current state is $y_j$ and the previous state was $y_{j-1}$; $0$ otherwise
\item $\phi(y_0) = 1$ if state of the first sentence in the sequence is $y_j$; $0$ otherwise
\end{itemize}

%% file: experiments.tex
\label{experiments}

In this Section we describe our data, baselines and experimental set-up.

\subsection{Datasets}
\label{data}

Our primary dataset consists of a collection of summaries (`Plot Overviews') of 300 English novels extracted from the `Literature Study Guides' section of SparkNotes~\footnote{\url{http://www.sparknotes.com/lit/}}. We pre-processed each of these summaries as described in Sec.~\ref{features}. Thereafter, we considered all pairs of characters that appeared together in at least five sentences in the respective summaries and arranged these sentences in order of appearance in the original summary.  We refer to these sequences of sentences as simply a sequence. This yielded a collection of 634 sequences consisting of a total of 5542 sentences. 

As noted in Sec.~\ref{model}, our semi-supervised framework is trained on fully and partially labeled sequences. For our experiments, we manually annotated a set of 100 sequences (consisting of a total of 792 annotated sentences). Out of these, 50 sequences (402 sentences) were fully annotated with a binary relationship state (for the two characters) for each sentence in the sequence. Continuous assignments of identical states were automatically collapsed into one to yield a shorter relationship sequence. We also partially annotated a set of another 50 sequences, which included annotating at least one sentence of the sequence with a binary relationship state. These 50 sequences consisted of about 390 sentences out of which 201 were annotated. 
(The dataset is available on the first author's webpage.)

For this work we considered summaries of novels instead of their complete text because we found summaries to be more precise and informative. Due to the inherent third-person narration style of summaries, they contain more explicit evidences about relationships. On the other hand, while processing novel texts directly one would have to infer these evidences from dialogues and subtle cues. While this is an interesting task in itself, we leave this exploration for future. 

We considered another dataset for evaluating our model. This dataset was collected independently by another set of authors using Amazon Mechanical Turk~\footnote{https://www.mturk.com/}. The annotators were shown summaries of novels and a list of characters appearing in the novel. They were then asked to choose pairs of characters and annotate if the relationship between them changed during the novel (binary annotations). They were also asked other questions, such as the overall nature of their relationship etc., which were not relevant for our problem. 
From this annotated dataset, 62 pairs of characters were present in our dataset. Out of these 62, the relationship was annotated as `changed' (positive class) for 20\% of the pairs. This dataset can be viewed as providing additional binary ground truth information about the sequences in our primary dataset and was used for evaluation only (and not for training). In this paper, we refer to this data as the AMT dataset (Citation will be provided in the final version of this paper).

\subsection{Baselines and Evaluation Measures}
\label{baseline}
Our primary baseline is an unstructured model that trains flat classifiers using the same content features as used by our framework but treats individual sentences of the sequences independently. We experimented with various models and report performances of Logistic Regression and Decision Tree. We compare our model with this baseline to test our hypothesis that the task of relationship sequence prediction is a structured problem, which benefits from remembering intra-novel history of relationship between characters. 

We also compare our framework, which employs a second order Markovian segmentation model, with an identical framework, which uses a similar segmentation model albeit with first order Markov assumption. This baseline is included to understand the importance of remembering a longer history of relationship between characters. Also, since a higher order model can look further back, it will discourage frequent changes in relationship status within the sequence more strongly.

For comparing performances of the various models we use two different performance measures. Our first measure accesses the goodness of the binary relationship state assignments for every sentence in the sequence using averaged Precisions (P), Recalls (R) and F1-measures (F) of the two states. The second evaluation measure, mimics a more practical scenario by evaluating from the perspective of the predicted relationship sequence, $\mathbf{r}$, instead of looking at individual sentences of the sequence. 
 It compares the `proximity' of the predicted relationship sequence to the ground truth sequence using Edit Distance 
 and reports mean Edit Distance (ED) over all test sequences. A better prediction model will be expected to have a smaller value for this Edit Distance based measure.

\begin{table}
\begin{center}
\begin{tabular}{|l|c|c|c|c|}\hline
\textbf{Model}	    & \textbf{P} & \textbf{R}   & \textbf{F}    & \textbf{ED} \\\hline\hline
J48                 & 54.98 & 51.83 & 50.05 & 0.98\\\hline
LR              	& 63.21 & 55.79 & 55.77 & 1.06 \\\hline
Order 1 Model       & 62.45 & 63.40 & 62.84 & 0.9 \\\hline
Order 2 Model   	& 66.49 & 65.97 & \textbf{66.22} & \textbf{0.66}\\\hline

\end{tabular}
\caption{Cross validation performances of various models. 
The second order model based framework outperforms the one that uses a first order model and the unstructured baselines LR and J48.}
\label{resultsTable}
\end{center}
\vspace{-0.1in}
\end{table} 

\begin{table}
\begin{center}
\begin{tabular}{|l|c|c|c|c|}\hline
\textbf{Model}	    & \textbf{P} & \textbf{R}   & \textbf{F} \\\hline\hline
J48              	& 49.46 & 49.17 & 40.95  \\\hline
LR              	& 52.81 & 54.33 & 44.33  \\\hline
Order 1 Model       & 53.54 & 54.83 & 44.78  \\\hline
Order 2 Model   	& 52.98 & 54.63 & \textbf{49.53}  \\\hline
\end{tabular}
\caption{Performance comparison on the AMT dataset. The second order model based framework outperforms the one that uses a first order model and the unstructured models LR and J48.}
\label{amtResultsTable}
\end{center}
\vspace{-0.1in}
\end{table}

\subsection{Evaluation on the primary dataset}
Table~\ref{resultsTable} compares 10-fold cross validation performances of our second order Semi-supervised Framework (Order 2 Model) with its first order counterpart (Order 1 Model) and two unstructured baselines: Decision Tree (J48) and Logistic Regression (LR). Since the performance of the semi-supervised frameworks depends on random initialization of the weights, the figures reported in the table are mean values over 100 random restarts. The number of relationship states, $|Y|$, was set to be 2 to correspond to the gold standard annotations. From the table we can see that the framework with the first order Markov model yields slightly better performance (higher averaged F-measure and lower mean Edit Distance) than the unstructured models (LR and J48). This hints at a need for modeling the information flow between sentences of the sequences. 
The further performance improvement with the second order model emphasizes this hypothesis and also demonstrates the benefit of remembering longer history of characters while making relationship judgments.

\subsection{Evaluation on the AMT dataset}
Table~\ref{amtResultsTable} compares performances of the various models on the AMT dataset using averaged Precision, Recall and F measures on the binary classification task of change prediction. The problem setting, input sequences format and the training procedure for these models is same as above. However, the models produce structured output (relationship sequences) that need to be converted to the binary output of change prediction task. We do this simply by predicting the positive class (change occurred) if the outputted relationship sequence contained at least one change. We can see that while the performance of the framework using the first order model is similar to that of the baseline LR, the second order model shows a considerable improvement in performance. A closer look at the F measures of the two classes (not reported due to space constraints) revealed that while the performance on the positive class was similar for all the models (except J48 which was lower), the performance on the negative class (no change) was much higher for the structured models (56.0 for LR and 57.4 and 67.8 for the First and Second order models respectively). This might have happened because the unstructured model looks at independent sentences and cannot incorporate historical evidence so it is least conservative in predicting a change, which might have resulted in low recall. The structured models on the other hand look at previous states and hence it can better learn to make coherent state predictions.

%% file: relatedWork.tex



\label{relatedWork}

The work presented in this paper is most closely related to the character-centric methods in computational narrative domain. \cite{Bamman:2013} presented two latent variable models for learning personas in summaries of films by incorporating events that affect the characters. In their subsequent work~\cite{Bamman:2014}, they automatically infer character personas in English Novels. Similarly \cite{Vargas:2014} extract character roles from unannotated folk tales based on their actions. Unlike our work, these approaches do not explicitly model the relationship between characters (though an end user can manually infer the nature of relationship between them based on the definitions of the persona types, if available).

On the other hand, previous work has also focused on constructing social networks from text, though the interpretation of links between people varies according to the goals. For example, \cite{Elson:2010} analyzed dialogue interactions between characters of British novels and serials to construct their social networks and used them to test literary theories about such communities. Their goals required them to model the edges of the network using `volume' of interactions rather than `nature' of relationships. \cite{Agarwal:2013} focused on analyzing the direction of social events to construct social network with unstructured text. They also do not model polarity of relationships. However, they emphasized the importance of using dynamic networks to understand their text. \cite{He:2013} presented a method to infer identities of speakers and use them to construct a social network showing familial or social relationships. Most of these approaches involving social networks used them to identify positive relationships between people. There has also been some interest in modeling both positive and negative relationships. \cite{Leskovec:2010} proposed signed social networks to model both kinds of relationships though they work is in general social media domain. More recently, \cite{Krishnan:2015} analyze movie scripts to construct a signed social network depicting formality of relationships between movie characters. Apart from domain of application, our work differs from these in the sense that we model `polarity' of relationships and do that in a dynamic fashion.


%% file: conclusion.tex
\label{conclusion}

In this paper we have addressed the problem of predicting dynamic relationships between pairs of characters in a narrative. We analyze summaries of novels to extract relationship trajectories that describe how the relationship evolved. Our semi-supervised framework uses a structured segmentation model that makes second-order Markov assumption to remember the `history' of characters and analyzes textual contents of summaries using rich semantic features that incorporate world knowledge. We demonstrate the utility of our model by comparing it with an unstructured model that treats individual sentences independently and also with a lower order model that remembers shorter history.


Our experiments demonstrate that using a higher order model helps in making better predictions. In future we would like to experiment with models with order higher than 2 and also with semi-Markov models.

Also, this work treats different character pairs from the same novel independently and does not attempt to understand the complete text of the narrative (Sec.~\ref{caseStudy}). In future, we would like to explore a more sophisticated model that exploits intra-novel dynamics while predicting relationships.